\documentclass[sigconf]{acmart}
\AtBeginDocument{%
  }

\usepackage{graphicx}
\usepackage{subcaption}
\usepackage{amsmath}
\usepackage{float}

\begin{document}

\title{\texorpdfstring{AR\textsuperscript{2}: Adversarial Reinforcement Learning for Abstract Reasoning in Large Language Models}{AR\^2: Adversarial Reinforcement Learning for Abstract Reasoning in Large Language Models}}


\author{Cheng-Kai Yeh}
\orcid{0009-0007-3857-6784}
\affiliation{%
    \department{Department of Mathematical Sciences}
    \institution{National Chengchi University}
  \city{Taipei}
  \country{Taiwan}}
\email{111701035@g.nccu.edu.tw}

\author{Hsing-Wang Lee}
\orcid{0009-0006-1033-2633}
\email{111703050@g.nccu.edu.tw}
\author{Chung-Hung Kuo}
\orcid{0009-0000-6917-8817}
\email{111703054@g.nccu.edu.tw}
\affiliation{%
  \department{Department of Computer Science}
  \institution{National Chengchi University}
  \city{Taipei}
  \country{Taiwan}
}

\author{Hen-Hsen Huang}
\orcid{0000-0001-9169-3081}
\affiliation{%
  \department{Institute of Information Science}
  \institution{Academia Sinica}
  \city{Taipei}
  \country{Taiwan}
}
\email{hhhuang@iis.sinica.edu.tw}

\renewcommand{\shortauthors}{Yeh et al.}

\begin{abstract}
Abstraction--the ability to recognize and distill essential computational patterns from complex problem statements--is a foundational skill in computer science, critical both for human problem-solvers and coding-oriented large language models (LLMs). 
Despite recent advances in training LLMs for code generation using reinforcement learning (RL), most existing approaches focus primarily on superficial pattern recognition, overlooking explicit training for abstraction. 
In this study, we propose AR\textsuperscript{2} (Adversarial Reinforcement Learning for Abstract Reasoning), a novel framework explicitly designed to enhance the abstraction abilities of LLMs. 
AR\textsuperscript{2} employs a teacher model to transform kernel problems into narrative-rich, challenging descriptions without changing their fundamental logic. 
Simultaneously, a student coding model is trained to solve these complex narrative problems by extracting their underlying computational kernels. 
Experimental results demonstrate that AR\textsuperscript{2} substantially improves the student model's accuracy on previously unseen, challenging programming tasks, underscoring abstraction as a key skill for enhancing LLM generalization. 
\end{abstract}




\maketitle
\section{Introduction}
Large language models (LLMs) have achieved remarkable progress in code generation, with some models rivaling experienced human programmers~\citep{el2025competitive,deepseekai2025deepseekr1incentivizingreasoningcapability,quan2025codeelobenchmarkingcompetitionlevelcode}. 
Reinforcement learning (RL)~\citep{10.5555/3312046,wang2025reinforcementlearningenhancedllms} has proven especially effective compared to supervised fine-tuning (SFT)~\citep{chu2025sftmemorizesrlgeneralizes}, as it enables reward functions to incorporate practical constraints such as compilability~\citep{dou2024stepcoder} or test-case correctness~\citep{shojaee2023execution}, with recent advances alleviating training cost concerns~\citep{deepseek-math}.

Despite these gains, most training still targets producing correct code from given inputs, relying heavily on pattern matching rather than deeper problem understanding.
This overlooks a fundamental skill in computer science: abstraction--the ability to distill core computational principles from complex, domain-specific narratives. 
\textbf{Abstraction} enables both humans and LLMs to recognize structural similarities, transfer solutions across contexts, and generalize beyond memorized patterns.

\begin{figure}[!b]
    \centering
    \fbox{
    \begin{minipage}{0.45\textwidth}
        \textbf{Complex Problem (Narrative-rich description):}\\[5pt]
        Consider a graph \(G(V,E)\) where each node \(v \in V\) represents an element from the array \texttt{nums}. There exists an edge \((u,v) \in E\) if \(u\) and \(v\) are consecutive elements in the array. Given an integer \(k\), find the number of connected subgraphs \(H\) of \(G\) such that the number of distinct nodes in \(H\) is exactly \(k\).
    \end{minipage}
    }

    \vspace{10pt}
    \(\downarrow\) \textbf{Abstraction} \(\downarrow\)

    \vspace{10pt}
    \fbox{
    \begin{minipage}{0.45\textwidth}
        \textbf{Kernel Problem (Simplified abstraction):}\\[5pt]
        Given an integer array \texttt{nums} and an integer \(k\), return the number of good subarrays of \texttt{nums}. A good subarray is one in which the number of distinct integers is exactly \(k\).
    \end{minipage}
    }
    \caption{Illustration of abstraction from a complex problem to its computational kernel.}
    \label{fig:abstraction_example}
\end{figure}

Consider the example in \autoref{fig:abstraction_example}. The narrative-rich graph formulation initially appears complex, but skilled programmers use abstraction to reveal its computational kernel, namely counting subarrays with exactly $k$ distinct integers, and reduce it to a familiar array problem.
This process exemplifies a core principle in computer science: complex tasks can often be reframed as simpler, equivalent formulations, much like how many phenomena yield to fundamental mathematical or physical explanations.

While prior work has used LLMs to generate puzzles for self-play~\citep{haluptzok2022language} or to produce purely code-based tasks~\citep{haluptzok2022generating}, our focus is on generating natural-language, narrative-rich problems that preserve computational equivalence. This setup explicitly trains models to recover the core idea from varied surface forms.

We propose AR\textsuperscript{2} (Adversarial Reinforcement Learning for Abstract Reasoning), a teacher–student framework where a teacher rewrites well-defined LeetCode problems into challenging narrative-rich forms, and a student learns to solve them by extracting their kernels. A key innovation is requiring computational equivalence, enabling reuse of the original test cases for direct, stable reward computation during RL training.
Experiments show that AR\textsuperscript{2}  substantially improves student performance on unseen competitive programming tasks, with strong generalization across languages and problem types.
Our contributions are threefold as follows.
\begin{itemize}
    \item Identify abstract reasoning as a key capability for LLMs.
    \item Propose AR\textsuperscript{2}, an adversarial RL framework that effectively improves abstraction skills across benchmarks.
    \item Introduce a novel computationally equivalent problem-generation setting, simplifying reward computation.
    \item The dataset is released as a resource for the research community.\footnote{\url{https://github.com/hhhuang/ARAR}}
\end{itemize}

\begin{figure}[!t]
    \centering
    \includegraphics[clip, trim=0.5cm 6cm 0cm 0cm,width=0.8\textwidth]{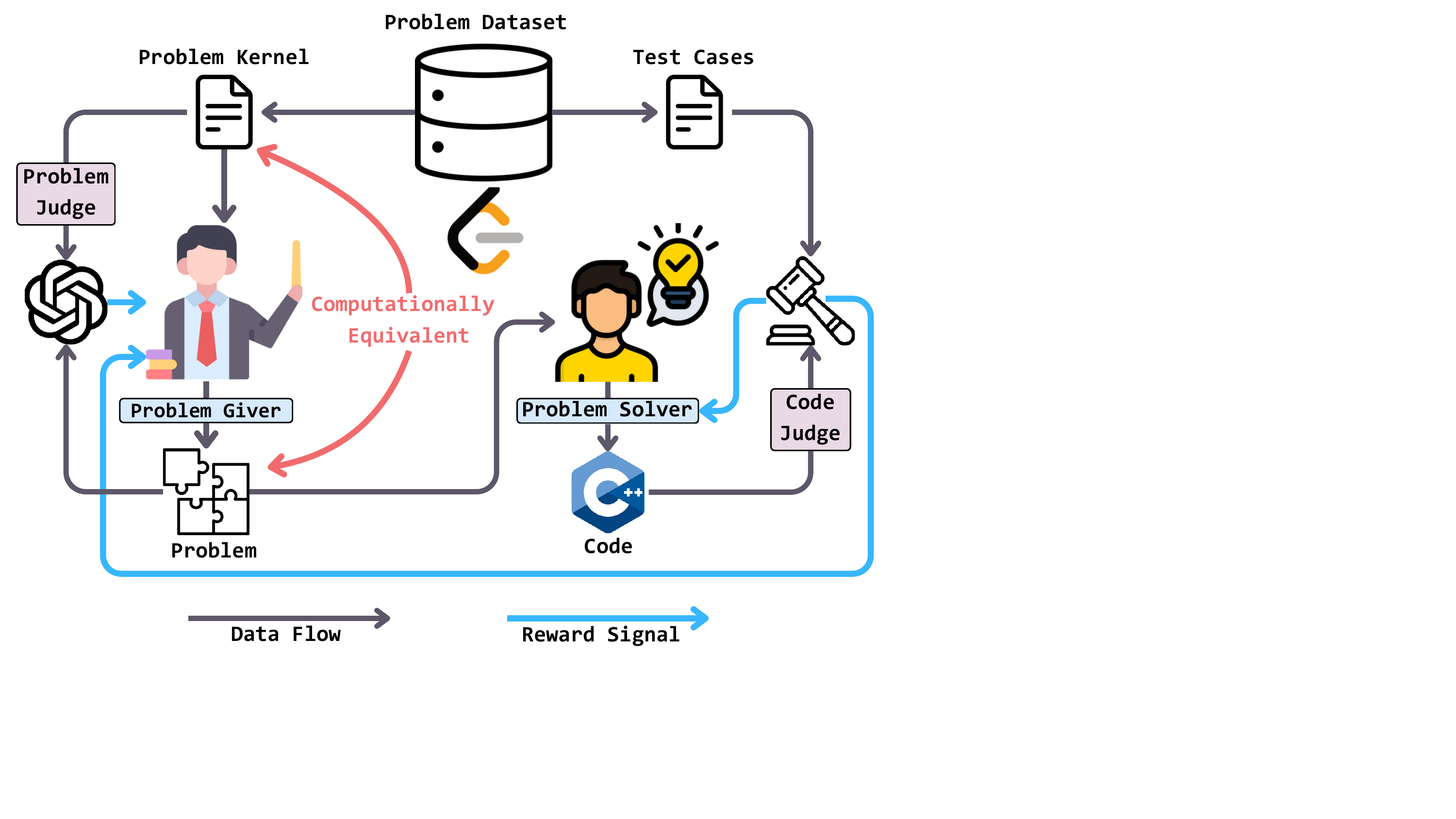}
    \caption{Overview of our AR\textsuperscript{2} framework for enhancing abstract reasoning in LLMs via adversarial reinforcement learning. A Problem Giver (teacher) rewrites a simple kernel problem into a computationally equivalent, narrative-rich version, preserving the original test cases for direct and efficient reward computation. 
    The Problem Solver (student) extracts the underlying abstraction and generates a solution, which is evaluated by the Code Judge against the original kernel's test cases. 
    Reward signals flow back to both models, driving the teacher to create more challenging yet faithful problems and the student to improve abstraction and generalization skills. 
    This computational-equivalence design enables stable, automated evaluation while fostering abstraction-oriented learning.}
    \label{fig:framework}
\end{figure}



\section{Methodology}
\label{sec:method}
\subsection{Framework Overview}

As illustrated in \autoref{fig:framework}, our AR\textsuperscript{2} framework employs an adversarial RL loop to jointly train two models—a teacher and a student—with the explicit goal of enhancing abstract reasoning:

\begin{itemize}
\item \textbf{Problem Giver (Teacher)} $\mathcal{G}$: Generates narrative-rich yet computationally equivalent versions of simple, clearly-defined computational tasks (referred to as \textit{kernel problems}). This computational equivalence ensures that the test cases associated with the original problems remain applicable to the newly generated narratives.
\item \textbf{Problem Solver (Student)} $\mathcal{S}$: Solves the more complex, narrative-rich problems by identifying and extracting their underlying computational abstraction, then producing correct algorithmic solutions.
\end{itemize}

A key novelty of our AR\textsuperscript{2} framework lies in the requirement for the teacher model to maintain computational equivalence between the original kernel problems $x$ and their narrative-rich revisions $y$. 
This crucial property allows the same original test cases to directly evaluate the correctness of the student's solutions. Consequently, our approach significantly simplifies and stabilizes the reward computation during adversarial training, providing a clear and robust learning signal.

Starting from base teacher $\mathcal{G}_0$ and student $\mathcal{S}_0$, we train them over $T$ alternating adversarial steps:
\begin{itemize}
    \item In the teacher phase, \(\mathcal{S}_{t-1}\) solves problems generated by \(\mathcal{G}_{t-1}\); its performance yields reward \(R_{\mathcal{G}}\), which updates \(\mathcal{G}_{t}\).
    \item In the student phase, \(\mathcal{G}_{t}\) generates computationally equivalent rewrites to train \(\mathcal{S}_{t}\).
\end{itemize}

The teacher's reward $\mathcal{R_G}$ combines five factors-complexity, diversity, equivalence, novelty, and ability to challenge the student, while the student's reward $\mathcal{R_S}$ covers format correctness, compilability, and accuracy.
This loop drives the teacher to generate increasingly diverse, challenging yet equivalent problems, and the student to sharpen its abstraction and problem-solving skills. 
As a result, AR\textsuperscript{2} fosters strong abstract reasoning and generalization to diverse computational tasks.

\subsection{Adversarial Reinforcement Learning}
To efficiently train our teacher ($\mathcal{G}$) and student ($\mathcal{S}$) models, we employ the Group Relative Policy Optimization (GRPO) algorithm~\citep{deepseek-math}, specifically chosen for its reduced GPU memory footprint. Unlike conventional policy optimization algorithms, GRPO avoids the requirement for a separate value model by sampling multiple candidate responses for each input prompt (problem) and computing a relative reward across these candidates.

Given our multi-dimensional reward structure—which evaluates generated outputs along several aspects—we adopt a modified GRPO variant proposed by \citet{liu2025understanding}, wherein the standardization term is omitted from the calculation of the group-relative advantage. Additionally, we remove the KL-divergence regularization term to further optimize GPU memory utilization.

Formally, GRPO operates by first sampling a group of $g$ candidate responses $\{o_i\}_{i=1}^g$ from a prior policy $\pi_{\text{old}}$. The policy parameters $\theta$ are then updated by maximizing the following objective:

{
\footnotesize
\begin{equation*}
\begin{split}
    \mathcal{J}_{\text{GRPO}}&(\pi_\theta) = \mathbb{E}_{\mathbf{q}\sim P(\mathcal{D}), \{\mathbf{o}_i\}_{i=1}^g \sim \pi_\text{old}(\cdot|\mathbf{q})} \frac{1}{g}\sum_{i=1}^g \frac{1}{|\mathbf{o}_i|} \\
    &\sum_{t=1}^{|\mathbf{o}_i|} \left\{ \min \left[ \frac{\pi_\theta(o_{i,t} | \mathbf{q}, \mathbf{o}_{i,<t})}{\pi_\text{old}(o_{i,t} | \mathbf{q}, \mathbf{o}_{i,<t})} \hat{A}_{i,t}, \operatorname{clip} \left( \frac{\pi_\theta(o_{i,t} | \mathbf{q}, \mathbf{o}_{i,<t})}{\pi_\text{old}(o_{i,t} | \mathbf{q}, \mathbf{o}_{i,<t})}, 1 - \epsilon, 1 + \epsilon \right)  \hat{A}_{i,t} \right] \right\},
\end{split}
\end{equation*}
}

\noindent where $\mathcal{D}$ denotes our dataset, $\hat{A}_{i,t}$ represents the unstandardized advantage value, and the clipping operator with hyperparameter $\epsilon$ constrains the update step size.

This adapted version of GRPO enables effective adversarial training between our teacher–student pair, providing a stable, computationally efficient method for enhancing abstract reasoning in LLMs.

\subsection{Reward of the Problem Solver}
\label{sec:method-solver}
The student model $\mathcal{S}$ is trained with GRPO to generate correct code solutions for the narrative-rich problems produced by the teacher model $\mathcal{G}$. Each candidate solution produced by $\mathcal{S}$ receives a scalar reward $R_{\mathcal{S}}$, composed of three distinct components:

\begin{enumerate}
\item \textbf{Solution Format Reward} $\bigl(R_{\mathrm{fmt}}\bigr)$:\\
We require $\mathcal{S}$'s responses to clearly distinguish intermediate reasoning from the final code solution. Specifically, intermediate reasoning must be enclosed within $\langle\texttt{think}\rangle \ldots \langle\texttt{think}\rangle$, and the final code must appear within $\langle\texttt{answer}\rangle \ldots  \langle\texttt{answer}\rangle$. The solution format reward is defined as:
\[
       R_{\mathrm{sfm}} = 
      \begin{cases}
          1, & \text{if the response meets the required format,}\\
          -1, & \text{otherwise.}
      \end{cases}
\]
This ensures the model explicitly learns to separate its reasoning steps from the executable solution.
\item \textbf{Compilability Reward} \(\bigl(R_{\mathrm{cmp}}\bigr)\):\\
For correctly formatted solutions, the code must compile successfully. The compilability reward is defined as:
\[
  R_{\mathrm{cmp}} =
  \begin{cases}
    2, & \text{if the formatted code compiles successfully,}\\
    -2, & \text{if formatting is correct but compilation fails,}\\
    0,  & \text{otherwise.}
  \end{cases}
\]
This encourages the student model to produce syntactically valid, executable code.

\item \textbf{Accuracy Reward} \(\bigl(R_{\mathrm{acc}}\bigr)\):\\
Once compiled successfully, the student's solution is evaluated against test cases associated with the original kernel problem. The accuracy reward is given by:
\[
  R_{\mathrm{acc}} =
  \begin{cases}
    3, & \text{if the solution passes all test cases (``Accepted''),}\\
    -3, & \text{if the solution compiles but fails any test case,}\\
    0,  & \text{otherwise.}
  \end{cases}
\]
This strongly incentivizes correct algorithmic solutions and penalizes partially correct ones.
\end{enumerate}

The combined reward $R_{\mathcal{S}} = R_{\mathrm{sfm}} + R_{\mathrm{cmp}} + R_{\mathrm{acc}}$ guides the student model toward clearly formatted, syntactically correct, and fully accurate solutions, ultimately enhancing its problem-solving and abstract reasoning capabilities.

\subsection{Reward of the Problem Giver}
\label{sec:method-teacher}
The teacher model $\mathcal{G}$ is trained via GRPO to generate semantically equivalent yet structurally diverse, narrative-rich variants of kernel problems. Each candidate problem revision receives a scalar reward $R_{\mathcal{G}}$, composed of five distinct components:

\begin{enumerate}
\item \textbf{Problem Format Reward} $\bigl(R_{\mathrm{pfm}}\bigr)$:\\
Similar to the solution format reward, each teacher-generated output must clearly separate its reasoning and final output using \texttt{<think>} and \texttt{<answer>} tags, respectively. 
Additionally, a valid problem must include a problem description, input and output formats, examples, and constraints—all within the $\texttt{<answer>}$ block. The reward is defined as:
\[       R_{\mathrm{pfm}} =
      \begin{cases}
        1, & \text{if the format is correct,} \\
        0, & \text{otherwise.}
      \end{cases}
\]
\item \textbf{Computational Equivalence Reward} \(\bigl(R_{\mathrm{eqv}}\bigr)\):\\
To ensure semantic fidelity, we verify computational equivalence between the revised problem and its original kernel using GPT-o3. The equivalence reward is defined as:
\[
  R_{\mathrm{eqv}} =
  \begin{cases}
    1, & \text{if } R_\mathrm{pfm} = 1 \text{ and computationally equivalent,} \\
    0, & \text{otherwise.}
  \end{cases}
\]

\item \textbf{Semantic Diversity Reward} \(\bigl(R_{\mathrm{dvg}}\bigr)\):\\
To encourage narrative-rich transformations, we measure divergence between the original kernel problem \(x\) and its revision \(y_t\) at iteration \(t\). Specifically, divergence is calculated using cosine similarity between their TF-IDF vector representations, denoted as \(s_{x,y_t}\). Thus, the divergence reward is defined as:
\[
  R_{\mathrm{dvg}} =
  \begin{cases}
    1 - s_{x,y_t}, & \text{if } R_\mathrm{eqv} = 1, \\
    0, & \text{otherwise.}
  \end{cases}
\]

\item \textbf{Iterative Novelty Reward} \(\bigl(R_{\mathrm{nvt}}\bigr)\):\\
To discourage repetitive or trivial edits, we compare the current revision \(y_t\) with the previous iteration's revision \(y_{t-1}\), again using cosine similarity \(s_{y_t,y_{t-1}}\) of TF-IDF vectors. The iterative novelty reward is defined as:
\[
  R_{\mathrm{nvt}} =
  \begin{cases}
    1 - s_{y_t, y_{t-1}}, & \text{if } R_\mathrm{eqv} = 1, \\
    0, & \text{otherwise.}
  \end{cases}
\]

\item \textbf{Adversarial Student-Failure Reward} \(\bigl(R_{\mathrm{adv}}\bigr)\):\\
After confirming computational equivalence, we test the student's ability to solve the generated problem and record its binary accuracy \(\mathrm{Acc} \in \{0, 1\}\). The adversarial reward thus incentivizes revisions that the student model fails to solve:
\[
  R_{\mathrm{adv}} =
  \begin{cases}
    1 - \mathrm{Acc}, & \text{if } R_\mathrm{eqv} = 1, \\
    0, & \text{otherwise.}
  \end{cases}
\]
\end{enumerate}

The combined teacher reward, $R_{\mathcal{G}} = R_{\mathrm{pfm}} + R_{\mathrm{eqv}} + R_{\mathrm{dvg}} + R_{\mathrm{nvt}} + R_{\mathrm{adv}}$, guides the teacher model towards producing complex, diverse, computationally equivalent problem revisions, pushing the student model toward stronger abstraction and problem-solving skills.

\section{Experiments}
\subsection{Experimental Setup}
We begin our experiments with a diverse set of 400 algorithm problems from LeetCode, covering various topics such as dynamic programming, graph theory, and greedy algorithms. To ensure dataset quality, we apply the following procedure to each problem:

\begin{enumerate}
\item Use \texttt{GPT-o3} to automatically generate a test-case generator capable of producing random, constraint-compliant inputs, including representative edge cases.
\item Use \texttt{GPT-o3} to produce a reference solution in C++, and verify correctness by compiling and executing it against the generated test cases.
\item Discard any problem for which the generated test-case generator or reference solution fails validation.
\end{enumerate}

After filtering, we retain a high-quality dataset comprising 300 validated problem–generator–solution triplets.

We first pre-train the teacher model $\mathcal{G}$ independently to ensure it reliably generates computationally equivalent problem revisions. Subsequently, we apply our AR\textsuperscript{2} framework (Section~\ref{sec:method}), alternately updating the teacher model $\mathcal{G}$ and the student model $\mathcal{S}$ in an iterative adversarial manner. In each iteration, we update $\mathcal{G}$ for 40 GRPO steps, followed by updating $\mathcal{S}$ for 100 GRPO steps. 
We employ Qwen 2.5 7B instruct~\citep{hui2024qwen2} as the base model for our teacher $\mathcal{G}$, and Qwen 2.5 7B coder instruct~\citep{qwen2.5} as the base model for our student $\mathcal{S}$. 
All training is conducted in bfloat16 precision with a maximum completion length of 2,048 tokens, no gradient clipping, and a fixed learning rate of $10^{-6}$.

We evaluate our framework using three widely recognized competitive programming benchmarks:
\begin{itemize}
\item \textbf{AtCoder:}
We construct a competitive-programming focused benchmark by selecting AtCoder problems from the LiveCodeBench (2409-2501) dataset~\citep{jain2024livecodebench} and evaluating the student model $\mathcal{S}$'s ability to generate correct, compilable C++ solutions. Each generated solution is evaluated against curated AtCoder test cases.

\item \textbf{HumanEval}~\citep{chen2021codex}:
This benchmark includes diverse programming tasks, ranging from basic language comprehension to more advanced reasoning and algorithmic skills, enabling us to evaluate the general coding proficiency of our student model $\mathcal{S}$.

\item \textbf{LiveCodeBench (2409-2501)}~\citep{jain2024livecodebench}:
Comprised of algorithm problems drawn from Codeforces, LeetCode, and AtCoder, this benchmark tests the student's ability to tackle challenging competitive programming tasks.  
\end{itemize}

For all benchmarks, we follow the standard evaluation protocol, reporting pass@1 scores computed from $n = 128$ candidate solutions per problem, sampled at a fixed temperature of 0.2.

\subsection{Results}
\begin{table}[!t]
    \centering
    \caption{Results on AtCoder (C++), HumanEval (C++), and LiveCodeBench (Python). 
    The base model is Qwen 2.5 7B Coder. 
    Following prior work, we report $\text{pass@1}$ scores.}
    \label{tab:results}
    \begin{tabular}{lcccc}
        \toprule
        Method  & AtCoder & HumanEval & LiveCodeBench  \\
        \midrule
        Base Model          & 18.283              
        & 85.318  
        & 18.660                                      \\
        Base Model w/ RL         & 19.107            
        & 87.752
        &  18.956                     \\
        Base Model w/ AR\textsuperscript{2}              & \textbf{20.322}           &  \textbf{88.571}           
        & \textbf{19.652}
        \\
        \bottomrule
    \end{tabular}
\end{table}

The results presented in Table~\ref{tab:results} 
demonstrate that our AR\textsuperscript{2} adversarially trained model substantially outperforms both the base model (Qwen 2.5 7B Coder) and the base model trained with standard reinforcement learning (RL) across multiple programming benchmarks. 
Our method consistently achieves strong performance gains, highlighting the significant impact of our adversarial training framework.
Remarkably, though trained on C++, the student also solves Python problems in LiveCodeBench, showing emergent cross-language reasoning.
This underscores that adversarial training effectively enhances abstraction and generalization capabilities, enabling robust knowledge transfer beyond the originally targeted programming language.

\subsection{Case Analysis}
The complex problem in Figure~\ref{fig:abstraction_example} was generated by our problem giver $\mathcal{G}$, transforming a simple array-based counting task into a graph-theoretic formulation while preserving the underlying computational kernel. 
The shift in terminology-from ``subarray'' to ``connected subgraph'' and from ``array elements'' to ``graph nodes''—creates a domain shift that may mislead solvers into considering unnecessary graph algorithms. 
Correctly solving the problem requires recognizing that, under this specific graph construction, connected subgraphs correspond exactly to contiguous subarrays.

From an abstraction-learning perspective, this example highlights the intended role of $\mathcal{G}$: forcing the problem solver $\mathcal{S}$ to strip away narrative and domain-specific wrapping to recover the computational kernel. 
Because computational equivalence is preserved, the original test cases remain valid, enabling direct and reliable reward computation. This ensures that improvements on narrative-rich problems reflect genuine gains in abstraction and reduction capabilities rather than superficial pattern matching.

\section{Conclusion}
In this work, we proposed AR\textsuperscript{2}, an adversarial RL framework for enhancing abstraction in LLMs through a teacher–student setup generating computationally equivalent yet challenging problems. 
AR\textsuperscript{2} achieved notable gains in abstraction and cross-language generalization across benchmarks; future work will extend this approach to broader domains.


\begin{acks}
We thank the meta-reviewer and reviewers for their valuable feedback, which greatly improved this work. 
This research was partially supported by the National Science and Technology Council (NSTC), Taiwan (Grant No. 112-2221-E-001-016-MY3), 
Academia Sinica (Grant No. 236d-1120205), and by the National Center for High-performance Computing (NCHC), National Applied Research Laboratories (NARLabs), and NSTC under the ``Trustworthy AI Dialog Engine (TAIDE)'' project.
\end{acks}


\section*{GenAI Usage Disclosure}
In this work, we employed AI assistance (specifically, large language models such as GPT-o3 and GPT-4.5) for the following purposes:

\begin{itemize}
\item Automatic generation and validation of test cases and reference solutions for initial dataset construction.
\item Verification of computational equivalence between generated narrative-rich problems and their kernel problems.
\item Assistance in refining and polishing the manuscript text.
\end{itemize}

The core ideas, methodologies, experimental analyses, and interpretation of results were entirely performed by the authors. AI tools were used exclusively to automate certain mechanical or verification tasks, thereby ensuring transparency and reproducibility.

\bibliographystyle{ACM-Reference-Format}
\bibliography{sample-base}

\appendix

\section{Hyper-parameter Setup}
Table~\ref{tab:settings} summarizes the hyper-parameter configurations used for training the baseline solver and the teacher–student models in our AR\textsuperscript{2} framework. 

\begin{table}[h]
    \centering
    \caption{Hyper-parameter configurations for the baseline solver, teacher, and student models.}
    \label{tab:settings}
    \begin{tabular}{lccc}
    \toprule
    Hyper-parameter & Baseline & Teacher & Student \\
    \midrule
    Beta & 0 & 0 & 0 \\
    Scale reward & False & False & False \\
    Temperature & 0.7 & 0.7 & 0.7 \\
    Number of training epochs & 1 & 1 & 1 \\
    Precision & BF16 & BF16 & BF16 \\
    Max completion length & 1,700 & 2,000 & 1,700 \\
    Per-device training batch size & 3 & 3 & 3 \\
    Number of generations & 24 & 21 & 24 \\
    \bottomrule
    \end{tabular}
\end{table}




\section{Adversarial Reward Dynamics}
Understanding the reward dynamics of teacher and student models provides insight into how AR\textsuperscript{2} fosters abstraction learning. These reward trends correspond directly to the reward functions defined in Section~\ref{sec:method-solver} and Section~\ref{sec:method-teacher}. 
By tracking the curves, we can observe the adversarial interaction: the teacher adapts to challenge the student, while the student gradually improves to meet these challenges. 
The convergence of both curves indicates a stable balance, which directly supports the generalization improvements reported in the main experiments.  

\begin{figure}[t]
    \centering
    \includegraphics[width=0.5\textwidth]{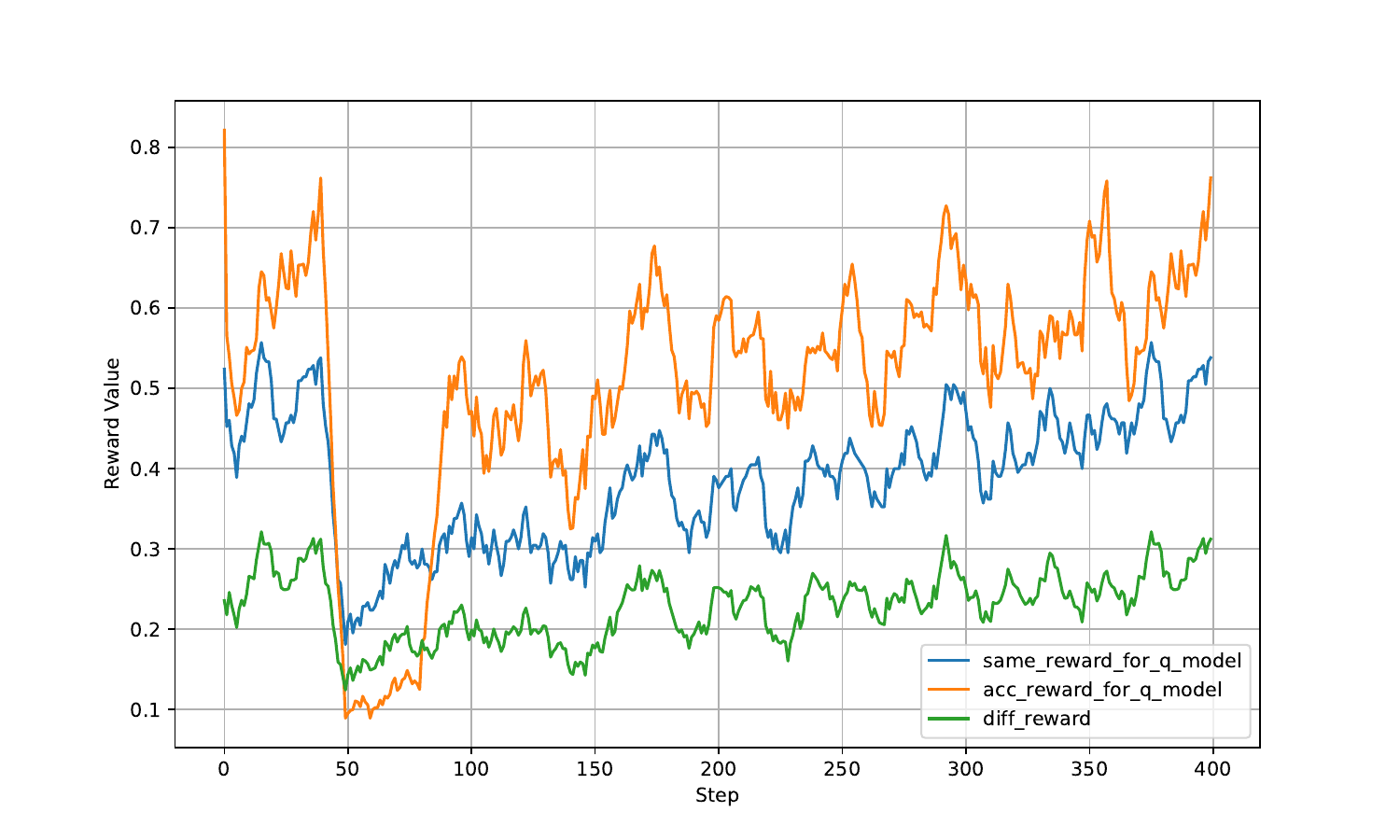}
    \caption{Teacher model reward curve during AR\textsuperscript{2} training. Convergence reflects the balance between generating challenging problems and student adaptation.}
    \label{fig:Q_reward}
\end{figure}

Figure~\ref{fig:Q_reward} shows the reward trend of the teacher model during AR\textsuperscript{2} training. At the beginning, the teacher obtains high rewards because the student has not yet adapted to the modified problems. As training progresses, the student becomes more capable, forcing the teacher to generate increasingly challenging transformations to maintain high rewards. Toward the end of training, the process converges to a balance between teacher and student.

\begin{figure}[t]
    \centering
    \includegraphics[width=0.5\textwidth]{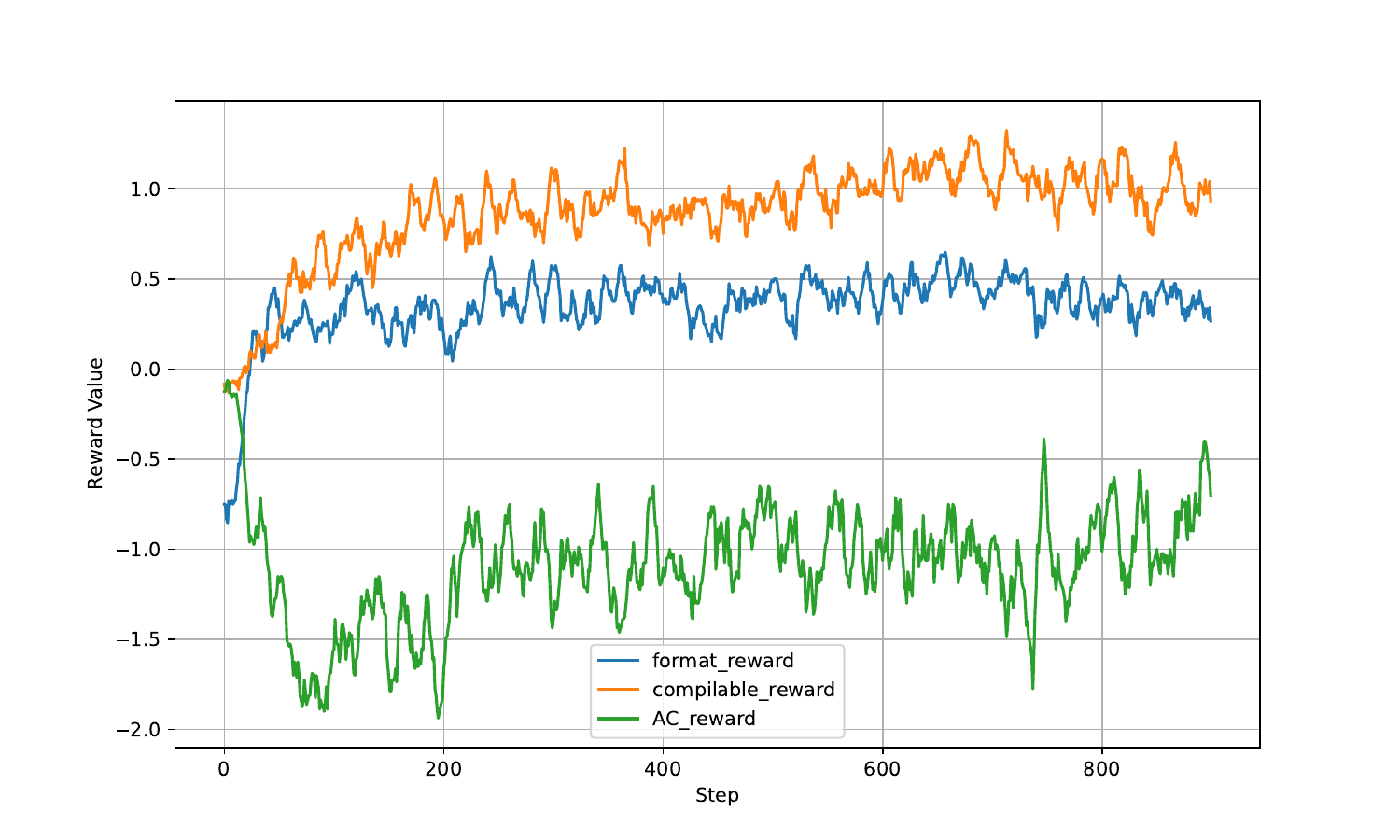}
    \caption{Student model reward curve during AR\textsuperscript{2} training. Convergence indicates successful adaptation to narrative-rich problems, supporting improved generalization.}
    \label{fig:A_reward}
\end{figure}

Figure~\ref{fig:A_reward} illustrates the reward curve of the student model. In the early iterations, the student struggles with the transformed problems, resulting in low accuracy-based rewards. After several training rounds, however, the student gradually adapts to the teacher's modifications and its performance converges.

Together, these curves capture the adversarial dynamics of AR\textsuperscript{2}.
The teacher must continually innovate to challenge the student, while the student incrementally strengthens its abstraction and problem-solving skills to catch up. The eventual convergence of both curves indicates that training stabilizes at a dynamic equilibrium, reflecting the intended balance between generating narrative-rich problems and solving them through abstraction. Unlike standard RL, where convergence may reflect memorization or overfitting, here it reflects genuine abstraction learning. Importantly, this equilibrium correlates with the observed performance gains on unseen benchmarks, showing that the adversarial balance not only stabilizes training but also drives improved generalization.

This analysis confirms the intended adversarial balance in AR\textsuperscript{2}, aligning the observed training dynamics with the theoretical motivation of fostering abstraction through teacher–student interaction.



\section{Dynamics of Problem Transformation}
\begin{figure}[t]
    \centering
    \fbox{
    \begin{minipage}{0.45\textwidth}
        \textbf{Kernel Problem :}\\[5pt]
        You are given an integer array \texttt{nums}. A good subsequence is defined as a subsequence of \texttt{nums} where the absolute difference between any two consecutive elements in the subsequence is exactly \(1\).
        Return the sum of all possible good subsequences of \texttt{nums}.
        Since the answer may be very large, return it modulo \(10^9 + 7\).
        Note that a subsequence of size \(1\) is considered good by definition.
    \end{minipage}
    }
    
\vspace{10pt}
\(\downarrow\) \textbf{Iteration 1} \(\downarrow\)
\vspace{10pt}

\fbox{
\begin{minipage}{0.45\textwidth}
    \textbf{Modified Problem:}\\[5pt]
    In a path graph where each node represents an element from the input list, we want to find the sum of all valid paths where the absolute difference between any two consecutive nodes is exactly $1$. All paths are considered modulo $10^9 + 7$.
\end{minipage}
}

\vspace{10pt}
\(\downarrow\) \textbf{After More Iteration} \(\downarrow\)
\vspace{10pt}

\fbox{
\begin{minipage}{0.45\textwidth}
    \textbf{Modified Problem:}\\[5pt]
    Given a list of nodes (integers), where each node can connect to nodes representing integers that are $1$ unit away (positive or negative), we need to count the total sum of all valid paths (subsequences) where each path has a difference of $1$ between its consecutive nodes. The final sum should be modulo $10^9 + 7$.

\end{minipage}
}
\caption{An example of teacher evolution: a kernel problem gradually transformed into graph-based formulations. While computationally equivalent, these rewrites increase representational complexity.}
\label{fig:Good_Transformation}
\end{figure}

\begin{figure}[t]
    \centering
    \fbox{
    \begin{minipage}{0.45\textwidth}
        \textbf{Kernel Problem :}\\[5pt]
With respect to a given puzzle string, a word is valid if both the following conditions are satisfied:
word contains the first letter of puzzle.
For each letter in word, that letter is in puzzle.
For example, if the puzzle is ``abcdefg'', then valid words are ``faced'', ``cabbage'', and ``baggage'', while
invalid words are ``beefed'' (does not include ``a'') and ``based'' (includes ``s'' which is not in the puzzle). Return an array answer, where answer[i] is the number of words in the given word list words that is valid with respect to the puzzle puzzles[i].
    
    \end{minipage}
    }

\vspace{10pt}
\(\downarrow\) \textbf{Wrong Transformation} \(\downarrow\)

\vspace{10pt}
\fbox{
\begin{minipage}{0.45\textwidth}
    \textbf{Modified Problem:}\\[5pt]
    Given a list of words and a list of puzzles where each word and puzzle consists of lowercase English letters, we need to count how many words in the list can form a valid word under a given puzzle. A word is valid if it includes the first character of the puzzle and contains all the characters from the puzzle.
\end{minipage}
}
\fbox{
\begin{minipage}{0.45\textwidth}
    \textbf{Reason:}\\[5pt]
    1. In the original problem, a word is valid if it (a) contains the first letter of the puzzle and (b) every letter in the word is found in the puzzle (i.e. the word is a subset of the puzzle's characters).
    
    2. In the rewritten problem, the validity condition is stated as ``includes the first character of the puzzle and contains all the characters from the puzzle.'' This erroneously implies that a valid word must include every single letter present in the puzzle, which is not the same as the original requirement.

Due to this discrepancy, the rewritten problem is not fully equivalent to the original problem.
\end{minipage}
}

\vspace{10pt}
\(\downarrow\) \textbf{Another Wrong Transformation} \(\downarrow\)
\vspace{10pt}

\fbox{
\begin{minipage}{0.45\textwidth}
    \textbf{Modified Problem:}\\[5pt]
    Design an algorithm to validate words against a set of puzzles based on the conditions specified in the problem. A valid word must start with the first letter of a puzzle and only contain characters present in that puzzle..

\end{minipage}
}
\fbox{
\begin{minipage}{0.45\textwidth}
    \textbf{Reason:}\\[5pt]
    In the rewritten version the condition ``must start with the first letter of a puzzle'' is used instead of ``contains the first letter of puzzle'' as stated originally. This subtle difference makes the rewritten problem not fully equivalent to the original.

\end{minipage}
}
\caption{Examples of flawed transformations. Semantic drift leads to stricter or altered validity conditions, breaking computational equivalence.}
\label{fig:Bad_Transformation}
\end{figure}

Figure~\ref{fig:Good_Transformation} shows an example of the dynamics of problem transformation. 
During training, the teacher model evolves its problem rewrites as it seeks to maintain high rewards. 
Early in training, the teacher often reformulates kernel problems into graph-like descriptions. Such transformations are natural (introducing nodes, edges, paths) but typically unnecessary: they increase representational complexity and diverge from the structure the solver expects. 
Because the teacher's reward function emphasizes textual variation (e.g., diversity and improvement rewards), the model tends to repeatedly rewrite prompts into graph-oriented language. This leads to \textit{semantic drift}: although the underlying combinatorial structure and solvability remain unchanged, the problem is expressed in a register that may unnecessarily complicate the task for the student model.

Not all transformations preserve computational equivalence. Sometimes the teacher introduces subtle logical errors that alter the problem definition. 
Figure~\ref{fig:Bad_Transformation} illustrates such cases, where semantic drift results in conditions that no longer match the original kernel.

This analysis highlights both the strengths and limitations of teacher-driven problem evolution. While the teacher can generate creative reformulations, reward design must carefully balance diversity with fidelity to avoid semantic drift and ensure true computational equivalence. 
As discussed in Section~\ref{sec:method-teacher}, this suggests that future work should incorporate stronger semantic checks or refined reward shaping to better align teacher transformations with the kernel's intended meaning.

Taken together, these observations help explain why carefully balanced problem evolution in AR\textsuperscript{2} contributes to the student model's improved generalization on competitive programming benchmarks.


\end{document}